%% file: paper.tex
\documentclass{article}

\PassOptionsToPackage{sort,numbers, compress}{natbib}

\usepackage[final]{nips_2016}

\usepackage[utf8]{inputenc} 
\usepackage[T1]{fontenc}    
\usepackage{hyperref}       
\usepackage{url}            
\usepackage{booktabs}       
\usepackage{amsfonts}       
\usepackage{nicefrac}       
\usepackage{microtype}      
\usepackage{color} 
\usepackage{xspace}  
\usepackage{graphicx}
\usepackage[export]{adjustbox}
\usepackage{algpseudocode,algorithm,algorithmicx}
\usepackage{amsmath}
\usepackage{caption}
\usepackage{subcaption}
\input{macros}

\title{Model-Free Episodic Control}

\author{
    Charles Blundell \\
    Google DeepMind \\
    \texttt{cblundell@google.com}
    \And
    Benigno Uria \\
    Google DeepMind \\
    \texttt{buria@google.com}
    \And
    Alexander Pritzel \\
    Google DeepMind \\
    \texttt{apritzel@google.com}
    \And
    Yazhe Li \\
    Google DeepMind \\
    \texttt{yazhe@google.com}
    \And
    Avraham Ruderman \\
    Google DeepMind \\
    \texttt{aruderman@google.com}
    \And
    Joel Z Leibo \\
    Google DeepMind \\
    \texttt{jzl@google.com}
    \And
    Jack Rae \\
    Google DeepMind \\
    \texttt{jwrae@google.com}
    \And
    Daan Wierstra \\
    Google DeepMind \\
    \texttt{wierstra@google.com}
    \And
    Demis Hassabis \\
    Google DeepMind \\
    \texttt{demishassabis@google.com}
}

\begin{document}	

\maketitle

\begin{abstract}
  \input{abstract}

\end{abstract}

\input{introduction}

\input{methods}
\input{experiments}

\input{discussion}

\newpage
\subsubsection*{Acknowledgements}
We are grateful to Dharshan Kumaran and Koray Kavukcuoglu for their detailed feedback on this manuscript.
We are indebted to Marcus Wainwright and Max Cant for generating the images in Figure~\ref{fig:screenshots}.
We would also like to thank Peter Dayan, Shane Legg, Ian Osband, Joel Veness, Tim Lillicrap, Theophane Weber, Remi Munos, Alvin Chua, Yori Zwols and many others at Google DeepMind for fruitful discussions.

\bibliography{references}
\bibliographystyle{plain}

\appendix

\input{appendix}

\end{document}

%% file: macros.tex
\newcommand{\todo}[1]{}
\renewcommand{\todo}[1]{{\color{red} TODO: {#1}}}

\newcommand{\noteAlex}[1]{}
\renewcommand{\noteAlex}[1]{{\color{blue}NOTE(Alex): {#1}}}

\newcommand{\noteAvi}[1]{}
\renewcommand{\noteAvi}[1]{{\color{blue}NOTE(Avi): {#1}}}

\newcommand{\noteBeni}[1]{}
\renewcommand{\noteBeni}[1]{{\color{blue}NOTE(Beni): {#1}}}

\newcommand{\noteCharles}[1]{}
\renewcommand{\noteCharles}[1]{{\color{blue}NOTE(Charles): {#1}}}

\newcommand{\noteYazhe}[1]{}
\renewcommand{\noteYazhe}[1]{{\color{blue}NOTE(Yazhe): {#1}}}

\newcommand*{\eg}{e.g.\@\xspace}
\newcommand*{\ie}{i.e.\@\xspace}
\newcommand*{\vs}{\textit{vs.\@\xspace}\@\xspace}

\newcommand{\expected}{\mathop{\mathbb{E}}}

\newcommand{\vx}{\mathbf{x}}
\newcommand{\vz}{\mathbf{z}}

\newcommand{\Qec}{\ensuremath{Q^{\text{EC}}}\xspace}

%% file: abstract.tex
State of the art deep reinforcement learning algorithms take many millions of interactions to attain human-level performance.
Humans, on the other hand, can very quickly exploit highly rewarding nuances of an environment upon first discovery.
In the brain, such rapid learning is thought to depend on the hippocampus and its capacity for episodic memory.
Here we investigate whether a simple model of hippocampal \emph{episodic control} can learn to solve difficult sequential decision-making tasks.
We demonstrate that it not only attains a highly rewarding strategy significantly faster than state-of-the-art deep reinforcement learning algorithms, but also achieves a higher overall reward on some of the more challenging domains.

%% file: introduction.tex
\section{Introduction}

Deep reinforcement learning has recently achieved notable successes in a variety of domains \cite{mnih2015human, silver2016mastering}.  However, it is very data inefficient.  For example, in the domain of Atari games \cite{bellemare2012arcade}, deep Reinforcement Learning (RL) systems typically require tens of millions of interactions with the game emulator, amounting to hundreds of hours of game play, to achieve human-level performance.
As pointed out by \cite{lake2016building}, humans learn to play these games much faster.  This paper addresses the question of how to emulate such fast learning abilities in a machine---without any domain-specific prior knowledge.

Current deep RL algorithms may happen upon, or be shown, highly rewarding sequences of actions. Unfortunately, due to their slow gradient-based updates of underlying policy or value functions, these algorithms require large numbers of steps to assimilate such information and translate it into policy improvement. Thus these algorithms lack the ability to rapidly latch onto successful strategies. Episodic control, introduced by \citep{ThirdWay}, is a complementary approach that can rapidly re-enact observed, successful policies. Episodic control records highly rewarding experiences and follows a policy that replays sequences of actions that previously yielded high returns.

In the brain, this form of very fast learning is critically supported by the hippocampus and related medial temporal lobe structures \cite{squire2004memory, andersen2006hippocampus}. For example, a rat's performance on a task requiring navigation to a hidden platform is impaired by lesions to these structures \cite{morris1982place, sutherland1983behavioural}. Hippocampal learning is thought to be instance-based \cite{marr1971simple, sutherland1989configural}, in contrast to the cortical system which represents generalised statistical summaries of the input distribution \cite{tulving1991long, mcclelland1995there, norman2003modeling}. The hippocampal system may be used to guide sequential decision-making by co-representing environment states with the returns achieved from the various possible actions. After such encoding, at a given probe state, the return associated to each possible action could be retrieved by pattern completion in the CA3 subregion  \cite{hopfield1982neural, mcnaughton1987hippocampal, treves1994computational, nakazawa2002requirement}. The final value achieved by a sequence of actions could quickly become associated with each of its component state-action pairs by the reverse-ordered replay of hippocampal place cell activations that occurs after a rewarding event \cite{foster2006reverse}. 

Humans and animals utilise multiple learning, memory, and decision systems each best suited to different settings \cite{squire1992memory, daw2005uncertainty}. For example, when an accurate model of the environment is available, and there are sufficient time and working memory resources, the best strategy is model-based planning associated with prefrontal cortex \cite{daw2005uncertainty}. However, when there is no time or no resources available for planning, the less compute-intensive immediate decision systems must be employed \cite{otto2013curse}. This presents a problem early on in the learning of a new environment as the model-free decision system will be even less accurate in this case since it has not yet had enough repeated experience to learn an accurate value function. In contrast, this is the situation where model-free episodic control may be most useful. Thus the argument for hippocampal involvement in model-free control parallels the argument for its involvement in model-based control. In both cases quick-to-learn instance-based control policies serve as a rough approximation while a slower more generalisable decision system is trained up \cite{ThirdWay}.

The domain of applicability of episodic control may be hopelessly limited by the complexity of the world.
In real environments the same exact situation is rarely, if ever, revisited.
In RL terms, repeated visits to the exactly the same state are also extremely rare.
Here we show that the commonly used Atari environments do not have this property. In fact, we show that the agents developed in this work re-encounter exactly the same Atari states between 10-60\% of the time. As expected, the episodic controller works well in such a setting. The key test for this approach is whether it can also work in more realistic environments where states are never repeated and generalisation over similar states is essential. Critically, we also show that our episodic control model still performs well in such (3D) environments where the same state is essentially never re-visited.

%% file: methods.tex
\section{The episodic controller}

In reinforcement learning~\cite[\eg][]{SuttonBarto}, an agent interacts with an environment through a sequence of states, $s_t \in \mathcal{S}$; actions, $a_t \in \mathcal{A}$; and rewards $r_{t+1} \in \mathbb{R}$.
Actions are determined by the agent's policy $\pi(a_t | s_t)$, a probability distribution over the action $a_t$.
The goal of the agent is to learn a policy that maximises the expected discounted return $R_{t}=\sum_{\tau=1}^{T-t}\gamma^{\tau-1} r_{t+\tau}$ where $T$ is the time step at which each episode ends, and $\gamma \in (0,1]$ the discount rate.
Upon executing an action $a_t$ the agent transitions from state $s_t$ to state $s_{t+1}$.

Environments with deterministic state transitions and rewards are common in daily experience. For example, in navigation, when you exit a room and then return back, you usually end up in the room where you started. This property of natural environments can be exploited by RL algorithms or brains.  However, most existing scalable deep RL algorithms (such as DQN~\cite{mnih2015human} and A3C~\cite{AsyncRL}) do not do so. They were designed with more general environments in mind. Thus, in principle, they could operate in regimes with high degrees of stochasticity in both transitions and rewards. This generality comes at the cost of longer learning times. DQN and A3C both attempt to find a policy with maximal \emph{expected} return. Evaluating the expected return requires many examples in order to get accurate estimates. Additionally, these algorithms are further slowed down by gradient descent learning, typically in lock-step with the rate at which actions are taken in the environment.

Given the ubiquity of such near-deterministic situations in the real world, it would be surprising if the brain did not employ specialised learning mechanisms to exploit this structure and thereby learn more quickly in such cases. The episodic controller model of hippocampal instance-based learning we propose here is just such a mechanism. It is a non-parametric model that rapidly records and replays the sequence of actions that so far yielded the highest return from a given start state. In its simplest form, it is a growing table, indexed by states and actions. By analogy with RL value functions, we denote this table $\Qec(s, a)$. Each entry contains the highest return ever obtained by taking action $a$ from state $s$.

The episodic control policy picks the action with the highest value in $\Qec$ for the given state.
At the end of each episode, $\Qec$ is updated according to the return received as follows:
\begin{eqnarray}
    \label{eq:qec}
    \Qec(s_t, a_t) &\leftarrow
    \begin{cases}
        R_t &\text{if $(s_t, a_t) \not\in\Qec$,} \\
        \max \left\{\Qec(s_t, a_t), R_t\right \} &\text{otherwise,}
    \end{cases}
\end{eqnarray}
where $R_t$ is the discounted return received after taking action $a_t$ in state $s_t$.
Note that \eqref{eq:qec} is not a general purpose RL learning update: since the stored value can never decrease,
it is not suited to rational action selection in stochastic environments.\footnote{Following a policy that picks the action with the highest $\Qec$ value would yield a strong risk seeking behaviour in stochastic environments.
It is also possible to, instead, remove the max operator and store $R_t$ directly.
This yields a less optimistic estimate and performed worse in preliminary experiments.}

Tabular RL methods suffer from two key deficiencies: firstly, for large problems they consume a large amount of memory, and secondly, they lack a way to  generalise across similar states.
To address the first problem, we limit the size of the table by removing the least recently updated entry once a maximum size has been reached.
Such forgetting of older, less frequently accessed memories also occurs in the brain \cite{hardt2013decay}.

In large scale RL problems (such as real life) novel states are common; the real world, in general, also has this property.
We address the problem of what to do in novel states and how to generalise values across common experiences by taking \Qec to be a non-parametric nearest-neighbours model.
Let us assume that the state space $\mathcal{S}$ is imbued with a metric distance.
For states that have never been visited, \Qec is approximated by averaging the value of the $k$ nearest states.
Thus if $s$ is a novel state then \Qec is estimated as
\begin{eqnarray}
    \label{eq:qecknn}
    \widehat{\Qec}(s,a) = 
    \begin{cases}
        \frac{1}{k} \sum_{i=1}^k \Qec(s^{(i)}, a) & \text{if $(s,a) \not\in \Qec$,} \\
        \Qec(s,a) & \text{otherwise,}
    \end{cases}
\end{eqnarray}
where $s^{(i)}$, $i=1,\ldots,k$ are the $k$ states with the smallest distance to state $s$.\footnote{
In practice, we implemented this by having one $k$NN buffer for each action $a\in\mathcal{A}$
and finding the $k$ closest states in each buffer to state $s$.}

Algorithm~\ref{alg:episodic} describes the most basic form of the model-free episodic control.
The algorithm has two phases.
First, the policy implied by \Qec is executed for a full episode, recording the rewards received at each step.
This is done by projecting each observation from the environment $o_t$ via an embedding function $\phi$ to
a state in an appropriate state space: $s_t = \phi(o_t)$, then selecting the action with the highest estimated return according to \Qec.
In the second phase, the rewards, actions and states from an episode are associated via a backward replay process into \Qec to improve the policy.
Interestingly, this backward replay process is a potential algorithmic instance of the awake reverse replay of hippocampal states shown by \cite{foster2006reverse}, although as yet, we are unaware of any experiments testing this interesting use of hippocampus.
\begin{algorithm}[h]
    \caption{Model-Free Episodic Control.
		\label{alg:episodic}}
	\begin{algorithmic}[1]
		\For{each episode}
            \For{$t = 1, 2, 3, \dots, T$}
                \State Receive observation $o_t$ from environment.
                \State Let $s_t = \phi(o_t)$.
                \State Estimate return for each action $a$ via \eqref{eq:qecknn}
                \State Let $a_t = \arg\max_a \widehat{\Qec}(s_t, a)$
                \State Take action $a_t$, receive reward $r_{t+1}$
		    \EndFor
		    \For{$t = T,T-1,\ldots,1$}
                \State Update $\Qec(s_t, a_t)$ using $R_t$ according to \eqref{eq:qec}.
		    \EndFor
		\EndFor
	\end{algorithmic}
\end{algorithm}

The episodic controller acts according to the returns recorded in \Qec, in an attempt to replay successful sequences of actions and recreate past successes.
The values stored in $\Qec(s, a)$ thus do not correspond to estimates of the expected return, rather they are estimates of the highest potential return for a given state and action, based upon the states, rewards and actions seen.
Computing and behaving according to such a value function is useful in regimes where exploitation is more important than exploration, and where there is relatively little noise in the environment.

\section{\label{sec:featex} Representations}

In the brain, the hippocampus operates on a representation which notably includes the output of the ventral stream \cite{suzuki1994perirhinal, brown2001recognition, leibo2015approximate}. Thus it is expected to generalise along the dimensions of that representation space~\cite{mcclelland1996considerations}.  Similarly, the feature mapping, $\phi$, can play a critical role in how our episodic control algorithm performs when it encounters novel states\footnote{One way to understand this is that this feature mapping $\phi$ determines the dynamic discretization of the state-space into Voronoi cells implied by the $k$-nearest neighbours algorithm underlying the episodic controller.}.

Whilst the original observation space could be used, this may not work in practice. For example, each frame in the environments we consider in Section~\ref{sec:experiments} would occupy around 28 KBytes of memory and would require more than 300 gigabytes of memory for our experiments.
Instead we consider two different embeddings of observations into a state space, $\phi$, each having quite distinctive properties in setting the inductive bias of the $\Qec$ estimator.

One way of decreasing memory and computation requirements is to utilise a random projection of the original observations into a smaller-dimensional space, \ie $\phi:x \to \mathbf{A}x$, where $\mathbf{A} \in \mathbb{R}^{F \times D}$ and $F \ll D$ where $D$ is the dimensionality of the observation.
For a random matrix $\mathbf{A}$ with entries drawn from a standard Gaussian, the Johnson-Lindenstrauss lemma implies that this transformation approximately preserves relative distances in the original space~\cite{johnson1984extensions}.
We expect this representation to be sufficient when small changes in the original observation space correspond to small changes in the underlying return. 

For some environments, many aspects of the observation space are irrelevant for value prediction. For example,  illumination and textured surfaces in 3D environments (\eg \textit{Labyrinth} in Section~\ref{sec:experiments}), and scrolling backgrounds in 2D environments (\eg \textit{River Raid} in Section~\ref{sec:experiments}) may often be irrelevant.
In these cases, small distances in the original observation space may not be correlated with small distances in action-value.
A feature extraction method capable of extracting a more abstract representation of the state space (e.g. 3D geometry or the position of sprites in the case of 2D video-games) could result in a more suitable distance calculation.
Abstract features can be obtained by using latent-variable probabilistic models.
Variational autoencoders (VAE; \cite{kingma2013auto,rezende2014stochastic}), further described in the supplementary material, have shown a great deal of promise across a wide range of unsupervised learning problems on images.
Interestingly, the latent representations learnt by VAEs in an unsupervised fashion can lie on well structured manifolds capturing salient factors of variation \citep[Figures 4(a) and (b)]{kingma2013auto}; \citep[Figure 3(b)]{rezende2014stochastic}.
In our experiments, we train the VAEs on frames from an agent acting randomly.
Using a different data source will yield different VAE features, and in principle features from one task can be used in another.
Furthermore, the distance metric for comparing embeddings could also be learnt.
We leave these two interesting extensions to future work.

%% file: experiments.tex
\section{Experimental results}
\label{sec:experiments}
We tested our algorithm on two environments: the Arcade Learning Environment~(Atari)~\cite{bellemare2012arcade}, and a first-person 3-dimensional environment called Labyrinth~\cite{AsyncRL}.
Videos of the trained agents are available online\footnote{\url{https://sites.google.com/site/episodiccontrol/}}.

The Arcade Learning Environment is a suite of arcade games originally developed for the Atari-2600 console. These games are relatively simple visually but require complex and precise policies to achieve high expected reward~\cite{mnih2015human}.

Labyrinth provides a more complex visual experience, but requires relatively simple policies \eg turning when in the presence of a particular visual cue.
The three Labyrinth environments are foraging tasks with appetitive, adversive and sparse appetitive reward structures, respectively.

For each environment, we tested the performance of the episodic controller using two embeddings of the observations $\phi$: (1) 64 random-projections of the pixel observations and (2) the 64 parameters of a Gaussian approximation to the posterior over the latent dimensions in a VAE.

For the experiments that use latent features from a VAE, a random policy was used for one million frames at the beginning of training, these one million observations were used to train the VAE. The episodic controller is started after these one million frames, and uses the features obtained from the VAE. Both mean and log-standard-deviation parameters were used as dimensions in the calculation of Euclidean distances. To account for the initial phase of training we displaced performance curves for agents that use VAE features by one million frames.

\subsection{Atari}
For the Atari experiments we considered a set of five games, namely: \textit{Ms. PAC-MAN}, \textit{Q*bert}, \textit{River Raid}, \textit{Frostbite}, and \textit{Space Invaders}. We compared our algorithm to the original DQN algorithm~\cite{mnih2015human}, to DQN with prioritised replay~\cite{PrioritizedReplay}, and to the asynchronous advantage actor-critic~\cite{AsyncRL}~(A3C), a state-of-the-art policy gradient method \footnote{We are forever indebted to Tom Schaul for the prioritised replay baseline and Andrei Rusu for the A3C baseline.}. Following~\cite{mnih2015human}, observations were rescaled to $84$ by $84$ pixels and converted to gray-scale. The Atari simulator produces 60 observations (frames) per second of game play. The agents interact with the environment 15 times per second, as actions are repeated 4 times to decrease the computational requirements. An hour of game play corresponds to approximately $200{,}000$ frames.

In the episodic controller, the size of each buffer (one per action) of state-value pairs was limited to one million entries. If the buffer is full and a new state-value pair has to be introduced, the least recently used state is discarded. The $k$-nearest-neighbour lookups used $k = 11$. The discount rate was set to $\gamma=1$.
Exploration is achieved by using an $\epsilon$-greedy policy with $\epsilon = 0.005$.
We found that higher exploration rates were not as beneficial, as more exploration makes exploiting what is known harder.
Note that previously published exploration rates (e.g., \cite{mnih2015human, AsyncRL}) are at least a factor of ten higher.
Thus interestingly, our method attains good performance on a wide range of domains with relatively little random exploration.

\begin{figure}[t]
	\includegraphics[width=0.98\textwidth]{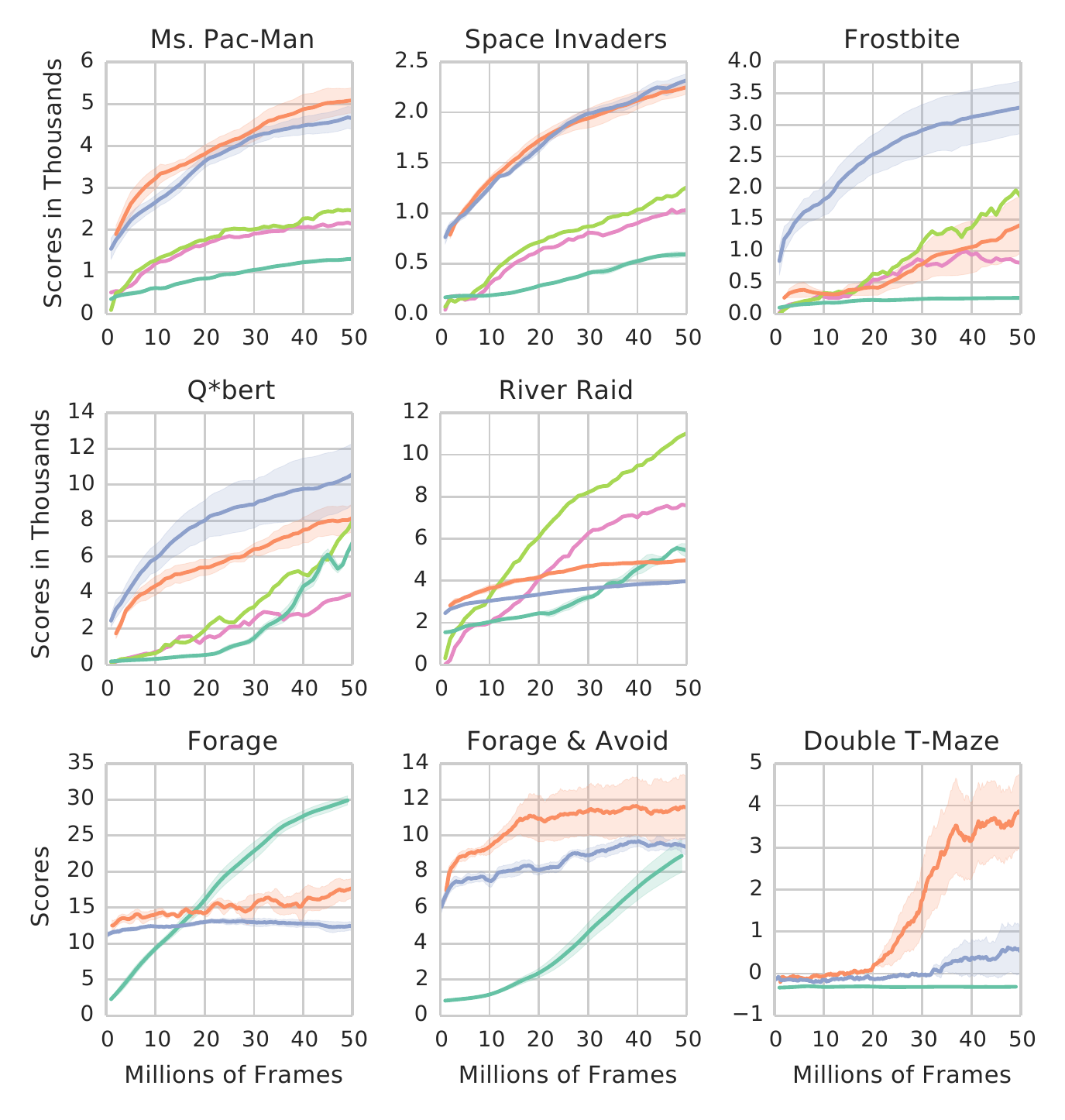}
	\includegraphics[width=1.0\textwidth]{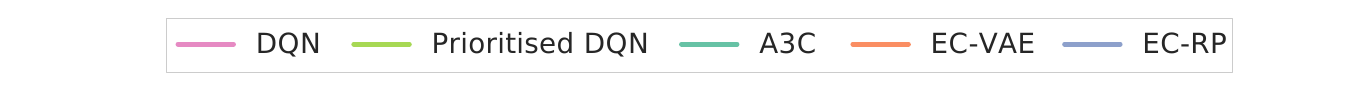}
	\caption{Average reward \vs number of frames (in millions) experienced for five Atari games and three Labyrinth environments. Dark curves show the mean of five runs (results from only one run were available for DQN baselines) initialised with different random number seeds. Light shading shows the standard error of the mean across runs. Episodic controllers (orange and blue curves) outperform parametric $Q$-function estimators (light green and pink curves) and A3C (dark green curve) in the initial phase of learning. VAE curves start after one million frames to account for their training using a random policy.}
	\label{fig:results}
\end{figure}

Results are shown in the top two rows of Figure~\ref{fig:results}. In terms of data efficiency the episodic controller outperformed all other algorithms during the initial learning phase of all games. On \textit{Q*bert} and \textit{River Raid}, the episodic controller is eventually overtaken by some of the parametric controllers (not shown in Figure~\ref{fig:results}). After an initial phase of fast learning the episodic controller was limited by the decrease in the relative amount of new experience that could be obtained in each episode as these become longer. In contrast the parametric controllers could utilise their non-local generalisation capabilities to handle the later stages of the games.

The two different embeddings (random projections and VAE) did not have a notable effect on the performance of the episodic control policies. Both representations proved more data efficient than the parametric policies. The only exception is \textit{Frostbite} where the VAE features perform noticeably worse. This may be due to the inability of a random policy to reach very far in the game, which results in a very poor training-set for the VAE.

Deep Q-networks and A3C exhibited a slow pace of policy improvement in Atari. For \textit{Frostbite} and \textit{Ms. PAC-MAN}, this has, sometimes, been attributed to naive exploration techniques~\cite{oh2015action,lake2016building}. Our results demonstrate that a simple exploration technique like  $\epsilon$-greedy can result in much faster policy improvements when combined with a system that is able to learn in a one-shot fashion.

The Atari environment has deterministic transitions and rewards. Each episode starts at one of thirty possible initial states. Therefore a sizeable percentage of states-action pairs are exactly matched in the buffers of $Q$-values: about $10\%$ for \textit{Frostbite}, $60\%$ for \textit{Q*bert}, $50\%$ for \textit{Ms. PAC-MAN}, $45\%$ for \textit{Space Invaders}, and $10\%$ for \textit{River Raid}. In the next section we report experiments on a set of more realistic environments where the same exact experience is seldom encountered twice.

\subsection{Labyrinth}

\begin{figure}
	\begin{tabular}{p{0.2cm} p{3.7cm} p{0.2cm} p{3.7cm} p{0.2cm} p{3.7cm}}
    \vspace{0pt}
	(a)		& 
	\vspace{0pt}
	\includegraphics[width=3.7cm]{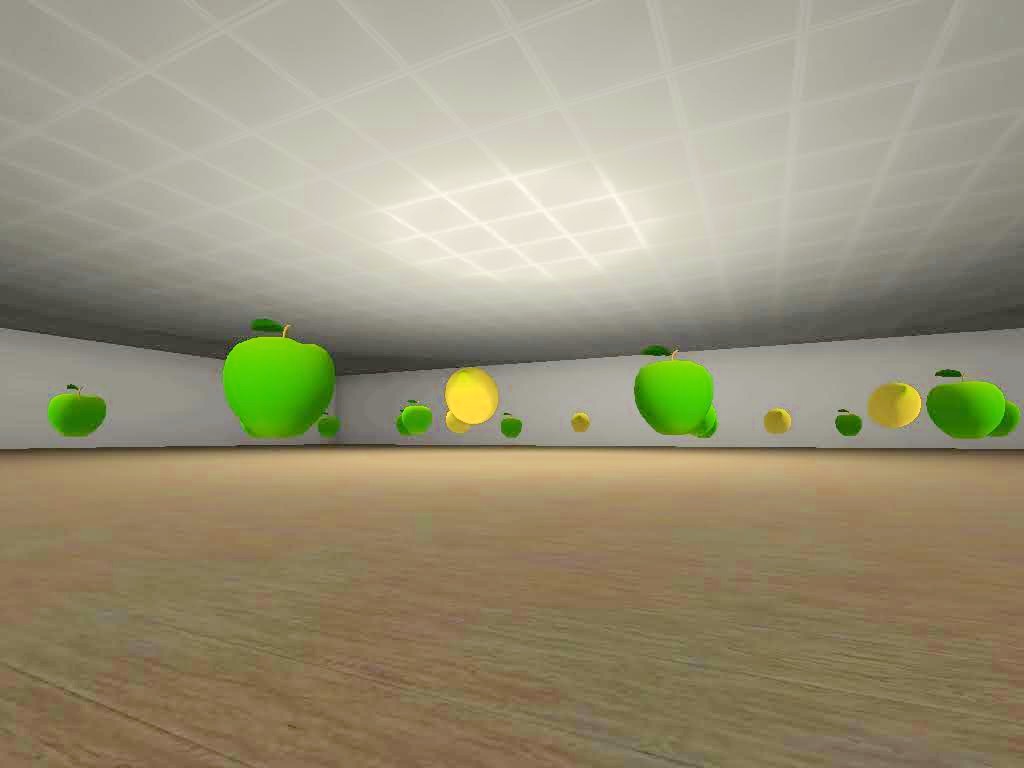} &
	\vspace{0pt}
	(b)		& 
	\vspace{0pt}
	\includegraphics[width=3.7cm]{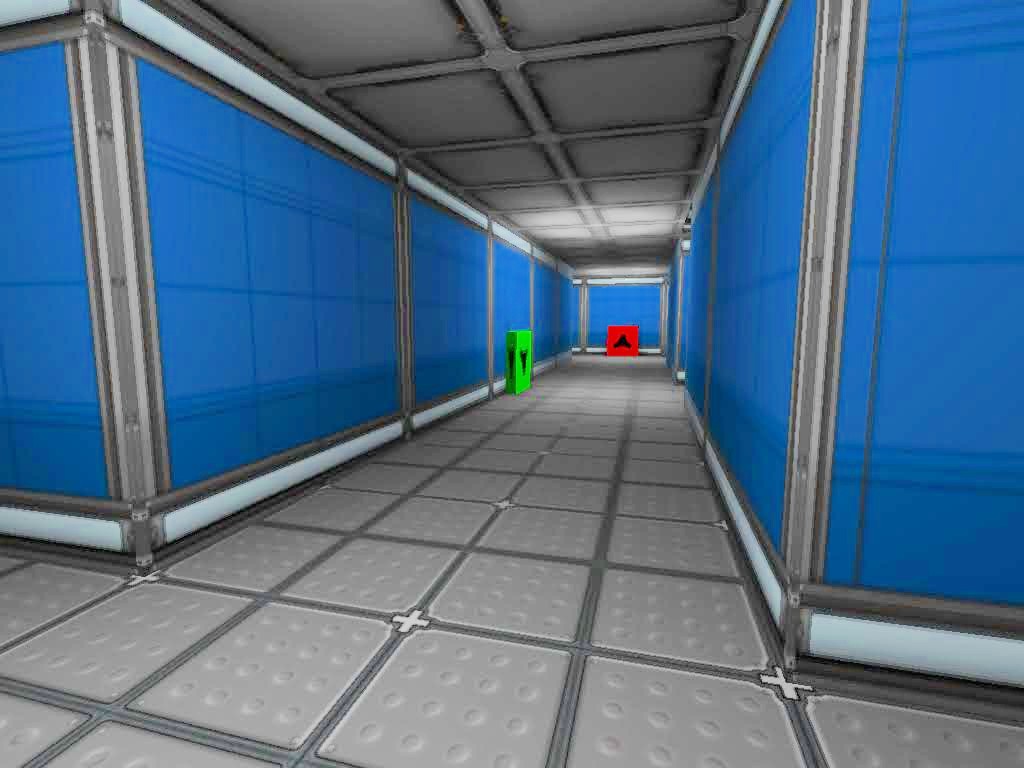} &
	\vspace{0pt}
	(c)		& 
	\vspace{0pt}
	\includegraphics[width=3.7cm]{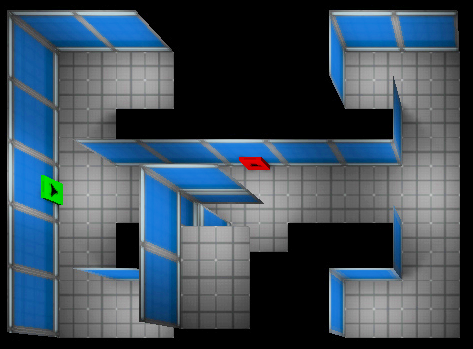}
	\end{tabular}
	\caption{
		High-resolution screenshots of the Labyrinth environments.
		\textbf{(a)} \textit{Forage and Avoid} showing the apples (positive rewards) and lemons (negative rewards).
		\textbf{(b)} \textit{Double T-maze} showing cues at the turning points.
		\textbf{(c)} Top view of a \textit{Double T-maze} configuration. The cues indicate the reward is located at the top left.
		}
	\label{fig:screenshots}
\end{figure}

The Labyrinth experiments involved three levels (screenshots are shown in Figure~\ref{fig:screenshots}).
The environment runs at 60 observations (frames) per simulated second of physical time. Observations are gray-scale images of $84$ by $84$ pixels. The agent interacts with the environment 15 times per second; actions are automatically repeated for 4 frames (to reduce computational requirements). The agent has eight different actions available to it (move-left, move-right, turn-left, turn-right, move-forward, move-backwards, move-forward and turn-left, move-forward and turn-right).
In the episodic controller, the size of each buffer (one per action) of state-value pairs was limited to one hundred thousand entries. When the buffer was full and a new state-value pair had to be introduced, the least recently used state was discarded. The $k$-nearest-neighbour lookups used $k = 50$. The discount rate was set to $\gamma=0.99$. Exploration is achieved by using an $\epsilon$-greedy policy with $\epsilon = 0.005$.
As a baseline, we used A3C \citep{AsyncRL}.
Labyrinth levels have deterministic transitions and rewards, but the initial location and facing direction are randomised, and the environment is much richer, being 3-dimensional. For this reason, unlike Atari, experiments on Labyrinth encounter very few exact matches in the buffers of $\Qec$-values; less than $0.1\%$ in all three levels.

Each level is progressively more difficult.
The first level, called \textit{Forage}, requires the agent to collect apples as quickly as possible by walking through them.
Each apple provides a reward of 1.
A simple policy of turning until an apple is seen and then moving towards it suffices here.
Figure~\ref{fig:results} shows that the episodic controller found an apple seeking policy very quickly.
Eventually A3C caught up, and final outperforms the episodic controller with a more efficient strategy for picking up apples.

The second level, called \textit{Forage and Avoid} involves collecting apples, which provide a reward of 1, while avoiding lemons which incur a reward of $-1$.
The level is depicted in Figure~\ref{fig:screenshots}(a).
This level requires only a slightly more complicated policy then \textit{Forage} (same policy plus avoid lemons) yet A3C took 
over 40 million steps to the same performance that episodic control attained in fewer than 3 million frames.

The third level, called \textit{Double-T-Maze}, requires the agent to walk in a maze with four ends (a map is shown in Figure~\ref{fig:screenshots}(c)) one of the ends contains an apple, while the other three contain lemons.
At each intersection the agent is presented with a colour cue that indicates the direction in which the apple is located (see Figure~\ref{fig:screenshots}(b)): left, if red, or right, if green.
If the agent walks through a lemon it incurs a reward of $-1$.
However, if it walks through the apple, it receives a reward of $1$, is teleported back to the starting position and the location of the apple is resampled. The duration of an episode is limited to 1 minute in which it can reach the apple multiple times if it solves the task fast enough.
\textit{Double-T-Maze} is a difficult RL problem: rewards are sparse.
In fact, A3C never achieved an expected reward above zero.
Due to the sparse reward nature of the \textit{Double T-Maze} level, A3C did not update the policy strongly enough in the few instances in which a reward is encountered through random diffusion in the state space.
In contrast, the episodic controller exhibited behaviour akin to one-shot learning on these instances, and was able to learn from the very few episodes that contain any rewards different from zero.
This allowed the episodic controller to observe between 20 and 30 million frames to learn a policy with positive expected reward, while the parametric policies never learnt a policy with expected reward higher than zero.
In this case, episodic control thrived in sparse reward environment as it rapidly latched onto an effective strategy.

\subsection{Effect of number of nearest neighbours on final score}

\begin{figure}
    \centering
    \begin{subfigure}[b]{0.4\textwidth}
        \includegraphics[width=\linewidth]{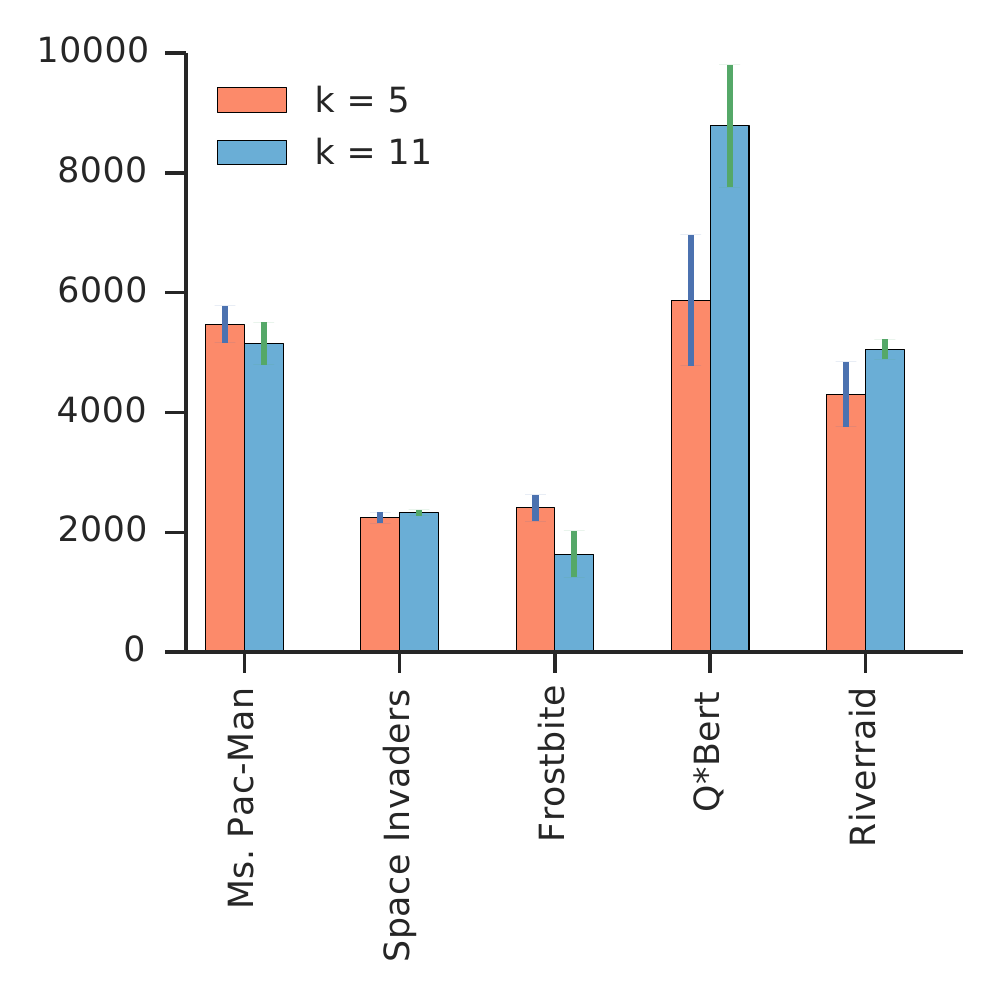}
        \caption{\label{fig:effKatari}Atari games.}
    \end{subfigure}
    \begin{subfigure}[b]{0.24\textwidth}
        \includegraphics[width=\linewidth]{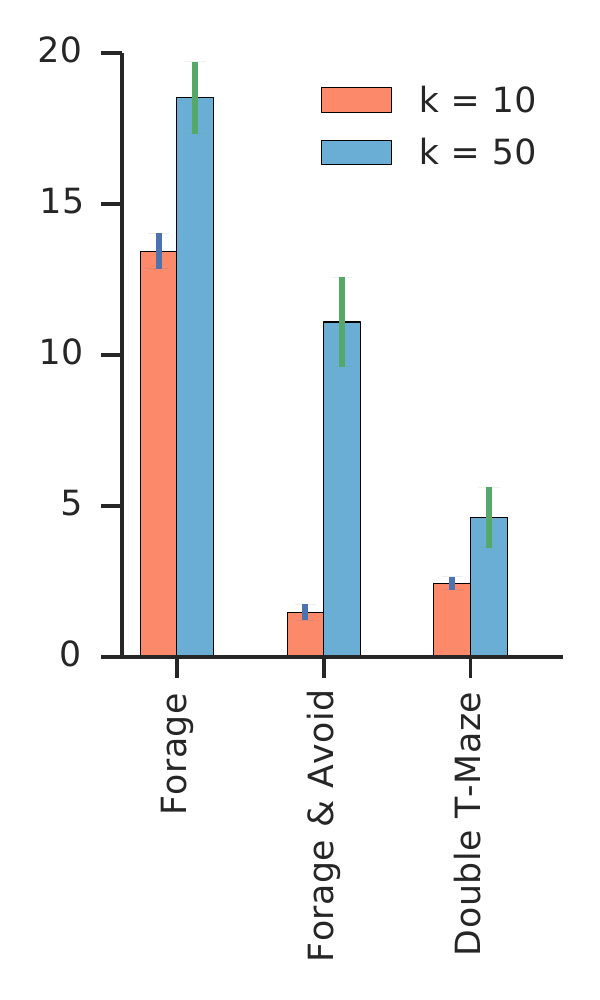}
        \caption{\label{fig:effKlab}Labyrinth levels.}
    \end{subfigure}
    \caption{Effect of number of neighbours, $k$, on on final score (y axis).}
\end{figure}

Finally, we compared the effect of varying $k$ (the number of nearest neighbours) on both Labyrinth and Atari tasks using
VAE features.
In our experiments above, we noticed that on Atari re-visiting the same state was common, and that random projections typically performed the same or
better than VAE features.
One further interesting feature is that the learnt VAEs on Atari games do not yield a higher score as the number of neighbours increases,
except on one game, \textit{Q*bert}, where VAEs perform reasonably well (see Figure~\ref{fig:effKatari}).
On Labyrinth levels, we observed that the VAEs outperformed random projections and the agent rarely encountered the same state more than once.
Interestingly for this case, Figure~\ref{fig:effKlab} shows that increasing the number of nearest neighbours has a significant effect on the
final performance of the agent in Labyrinth levels.
This strongly suggests that VAE features provide the episodic control agent with generalisation in Labyrinth.

%% file: discussion.tex
\section{Discussion}

This work tackles a critical deficiency in current reinforcement learning systems, namely their inability to learn in a one-shot fashion. We have presented a fast-learning system based on non-parametric memorisation of experience.  We showed that it can learn good policies faster than parametric function approximators. However, it may be overtaken by them at later stages of training.  It is our hope that these ideas will find application in practical systems, and result in data-efficient model-free methods. These results also provide support for the hypothesis that episodic control could be used by the brain, especially in the early stages of learning in a new environment. Note also that there are situations in which the episodic controller is always expected to outperform. For example, when hiding food for later consumption, some birds (e.g., scrub jays) are better off remembering their hiding spot exactly than searching according to a distribution of likely locations \cite{clayton1998episodic}. These considerations support models in which the brain uses multiple control systems and an arbitration mechanism to determine which to act according to at each point in time \cite{daw2005uncertainty, ThirdWay}.

We have referred to this approach as model-free episodic control to distinguish it from model-based episodic planning. We conjecture that both such strategies may be used by the brain in addition to the better-known habitual and goal-directed systems associated with dorsolateral striatum and prefrontal cortex respectively \cite{daw2005uncertainty}. The tentative picture to emerge from this work is one in which the amount of time and working memory resources available for decision making is a key determiner of which control strategies are available. When decisions must be made quickly, planning-based approaches are simply not an option. In such cases, the only choice is between the habitual model-free system and the episodic model-free system. When decisions are not so rushed, the planning-based approaches become available and the brain must then arbitrate between planning using semantic (neocortical) information or episodic (hippocampal) information. In both timing regimes, the key determiner of whether to use episodic information or not is how much uncertainty remains in the estimates provided by the slower-to-learn system. This prediction agrees with those of \cite{daw2005uncertainty, ThirdWay} with respect to the statistical trade-offs between systems. It builds on their work by highlighting the potential impact of  rushed decisions and insufficient working memory resources in accord with \cite{otto2013curse}. These ideas could be tested experimentally by manipulations of decision timing or working memory, perhaps by orthogonal tasks, and fast measurements of coherence between medial temporal lobe and output structures under different statistical conditions.

%% file: appendix.tex
\section{Variational autoencoders for representation learning}

\begin{figure}
	\centering
	\adjincludegraphics[height=5cm,trim={0 {.55\height} {.65\width} 0},clip]{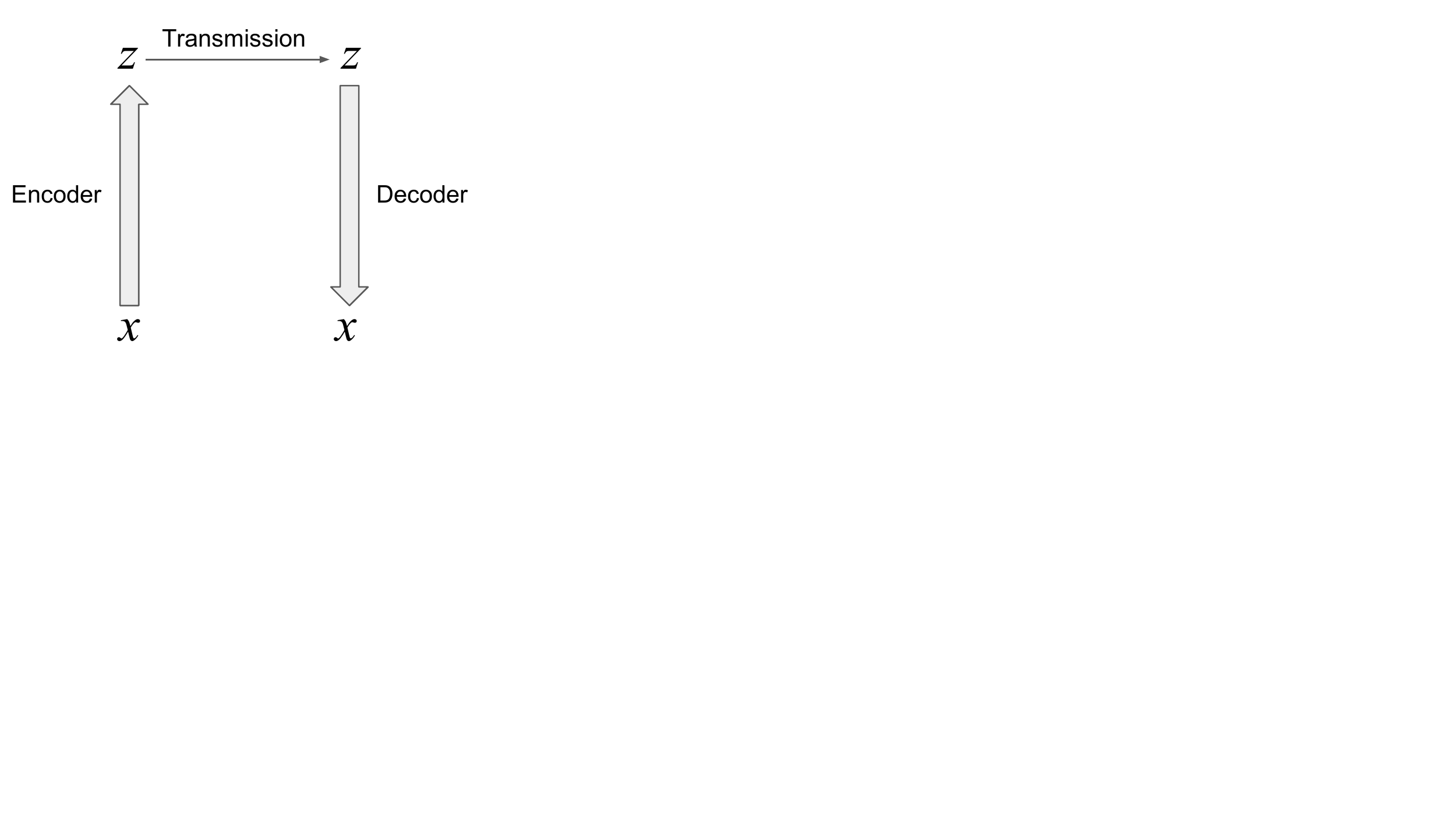}
	\caption{Diagram of a variational autoencoder.}
	\label{fig:vae}
\end{figure}

Variational autoencoders (VAE;~\cite{rezende2014stochastic,kingma2013auto}) are latent-variable probabilistic models inspired by compression theory. A VAE (shown in Figure~\ref{fig:vae}) is composed of two artificial neural networks: the encoder, which takes observations and maps them into messages; and a decoder, that receives messages and approximately recovers the observations. VAEs are designed to minimise the cost of transmitting observations from the encoder to the decoder through the communication channel. In order to minimise the transmission cost, a VAE must learn to capture the statistics of the distribution of observations~\cite[\eg][]{mackay2003information}. For our representation learning purposes, we use the encoder network as our feature mapping, $\phi$. for several data sets, representations learned by a VAE encoder have been shown to capture the independent factors of variation in the underlying generative process of the data~\cite{kingma2014semi}.

In more detail, the encoder receives an observation, $\vx$, and outputs the parameter-values for a distribution of messages, $q(z | x=\vx)$. The communication channel determines the cost of a message by a prior distribution over messages $p(z)$. The decoder receives a message, $\vz$, drawn at random from $q(z | x=\vx)$ and decodes it by outputting the parameters of a distribution over observations $p(x | z=\vz)$. VAEs are trained to minimise cost of exactly recovering the original observation, given by the sum of expected communication cost $KL\left( q(z| \vx)~||~p(z) \right)$ and expected correction cost $\expected \left[ p(x=\vx | z) \right]$.

In all our experiments, $x \in \mathbb{R}^{7056}$ (84 by 84 gray-scale pixels, with range [0, 1]), and $z \in \mathbb{R}^{32}$. We chose distributions $q(z | x)$, $p(z)$,  and $p(x|z)$ to be Gaussians with diagonal covariance matrices. In all experiments the encoder network  has four convolutional~\cite{lecun1998gradient} layers using \{32, 32, 64, 64\} kernels respectively, kernel sizes \{4, 5, 5, 4\}, kernel strides \{2, 2, 2, 2\} , no padding, and ReLU~\cite{nair2010rectified} non-linearity. The convolutional layer are followed by a fully connected layer of 512 ReLU units, from which a linear layer outputs the means and log-standard-deviations of the approximate posterior $q(\vz | \vx)$. The decoder is setup mirroring the encoder, with a fully connected layer of 512 ReLU units followed by four reverse convolutions~\cite{dosovitskiy2015learning} with \{64, 64, 32, 32\} kernels respectively, kernel sizes \{4, 5, 5, 4\}, kernel strides \{2, 2, 2, 2\}, no padding, followed by a reverse convolution with two output kernels --one for the mean and one for the log-standard-deviation of $p(x | z)$. The standard deviation of each dimension in $p(x|z)$ is not set to 0.05 if the value output by the network is smaller. The VAEs were trained to model a million observations obtained by executing a random policy on each environment. The parameters of the VAEs were optimised by running $400{,}000$ steps of stochastic-gradient descent using the RmsProp optimiser~\cite{tieleman2012lecture}, step size of $1\mathrm{e}{-5}$, and minibatches of size 100.

